\documentclass[conference]{IEEEtran}
\IEEEoverridecommandlockouts

\usepackage{cite}
\usepackage{amsmath,amssymb,amsfonts}
\usepackage{algorithmic}
\usepackage[ruled,vlined]{algorithm2e}
\usepackage{graphicx}
\usepackage{textcomp}
\usepackage{booktabs} 
\usepackage{hyperref}
\usepackage{graphicx}
\DeclareFontFamily{OT1}{pzc}{}
\DeclareFontShape{OT1}{pzc}{m}{it}{<-> s * [1.10] pzcmi7t}{}
\DeclareMathAlphabet{\mathpzc}{OT1}{pzc}{m}{it}
\usepackage{subcaption}
\usepackage[margin=1in]{geometry}
\usepackage{xcolor}
\def\BibTeX{{\rm B\kern-.05em{\sc i\kern-.025em b}\kern-.08em
    T\kern-.1667em\lower.7ex\hbox{E}\kern-.125emX}}
\begin{document}

\title{From Black-Box Tuning to Guided Optimization via Hyperparameters Interaction Analysis}



\author{\IEEEauthorblockN{1\textsuperscript{st} Moncef Garouani}
\IEEEauthorblockA{\textit{IRIT, UMR5505 CNRS} \\
\textit{ Université Toulouse Capitole}\\
Toulouse, France \\
moncef.garouani@irit.fr}
\and

\IEEEauthorblockN{2\textsuperscript{nd} Ayah Barhrhouj}
\IEEEauthorblockA{\textit{LIS, UMR 7020 CNRS} \\
\textit{Aix-Marseille University}\\
Marseille, France \\
ayah.barhrhouj@univ-amu.fr}
}

\maketitle

\begin{abstract}
Hyperparameters tuning is a fundamental, yet computationally expensive, step in optimizing machine learning models. Beyond optimization, understanding the relative importance and interaction of hyperparameters is critical to efficient model development. In this paper, we introduce \textbf{MetaSHAP}, a scalable semi-automated eXplainable AI\,(XAI) method, that uses meta-learning and Shapley values analysis to provide actionable and dataset-aware tuning insights. MetaSHAP operates over a vast benchmark of over 09 millions evaluated machine learning pipelines, allowing it to produce interpretable importance scores and actionable tuning insights that reveal \textit{how much} each hyperparameter matters, \textit{how it interacts} with others and \textit{in which value ranges} its influence is concentrated. For a given algorithm and dataset, MetaSHAP learns a surrogate performance model from historical configurations, computes hyperparameters interactions using SHAP-based analysis, and derives interpretable tuning ranges from the most influential hyperparameters. This allows practitioners not only to prioritize which hyperparameters to tune, but also to understand their directionality and interactions. We empirically validate MetaSHAP on a diverse benchmark of 164 classification datasets and 14 classifiers, demonstrating that it produces reliable importance rankings and competitive performance when used to guide Bayesian optimization. 
\end{abstract}


\section{Introduction}
\label{sec:introduction}

Hyperparameters optimization~(HPO) is a critical step in the machine learning~(ML) pipeline, as the choice of hyperparameters values can significantly influence the performance of the model. In practice, automated HPO strategies such as Bayesian Optimization~(BO)~\cite{BO}, Tree-structured Parzen Estimators~(TPE)~\cite{TPE}, or evolutionary algorithms~\cite{tpot} are employed to efficiently explore the hyperparameters space. While effective in discovering high-performing configurations, these methods typically conduct a blind exploration of the search space, without any prior knowledge of the relative importance of individual hyperparameters or their interactions, and provide little insight into why certain configurations perform better than others. This lack of guidance not only limits interpretability but also leads to inefficient search processes that significantly increase computational time, especially on large-scale datasets~\cite{mgbo}.

Moreover, existing HPO methods operate in a \emph{local} setting: they focus on optimizing hyperparameters for a specific algorithm–dataset pair, using performance signals solely from that task within a predefined search space. Although effective, local HPO discards valuable prior knowledge from previously solved tasks, and insights derived in this context often lack generalizability. In contrast, recent advances in meta-learning and automated ML (AutoML) have introduced the concept of \emph{global} HPO, where insights about algorithms and their hyperparameters behavior from previously solved tasks are transferred across datasets~\cite{meta-learning}. Global HPO facilitates knowledge reuse, reduces search costs, and allows algorithms or configurations recommendations. However, the challenge of generating meaningful, guided search spaces in global HPO remains largely open.

Recent work such as Auto-sklearn\,2~\cite{AutoSklearn2} has proposed initializing Bayesian optimization with recommendations derived from meta-learning, in order to identify promising starting configurations. However, this approach only supports the initial phase of the optimization process and does not fully exploit the actionable knowledge that can be extracted from prior experiments via meta-learning\,\cite{mtl_cef}. More recently, other studies have investigated how to assess the importance of hyperparameters that should be prioritized for tuning in a given algorithm–dataset setting. For instance, Meta-Guided BO~\cite{mgbo}, a framework that leverages functional ANOVA analysis~\cite{fanova}, has been proposed to identify which hyperparameters should be tuned within a Bayesian optimization process. Similarly, HyperSHAP~\cite{HyperSHAP} adapts Shapley value theory to compute the local importance of hyperparameters within HPO runs. While these approaches provide valuable insights to kick-start the optimization, they do not offer actionable guidance on how to navigate the hyperparameter search space itself. In practice, however, practitioners often face critical questions such as: Given a new dataset, \textit{which hyperparameters are likely to be important? How should the search space be prioritized or pruned? In which direction should each hyperparameter be tuned?}

To address these challenges, we introduce MetaSHAP, a novel framework that bridges meta-learning, Game theory, and hyperparameters interactions analysis. MetaSHAP leverages a large-scale knowledge base of over 09 million evaluated ML pipelines to perform dataset-aware, Shapley-based hyperparameters interactions analysis. Given a new dataset and a learning algorithm, MetaSHAP retrieves similar datasets based on meta-feature representations, fits surrogate performance models, and aggregates Shapley values to identify which hyperparameters and interactions among them are likely to influence performance. Our contributions are threefold:
\begin{itemize}
\item We propose a formalization of global, dataset-aware informed hyperparameters tunability grounded in Game theory.
\item We introduce a scalable architecture that combines meta-learning-based retrieval, surrogate modeling, and SHAP-based analysis over a large benchmark of more than 09 millions ML pipelines.
\item We empirically validate MetaSHAP, showing that it provides efficient, actionable hyperparameters tunability insights across algorithms-datasets pairs.
\end{itemize}

The remainder of this paper is organized as follows: Section~\ref{sec:background} provides the theoretical background. Section~\ref{sec:related_work} reviews relevant literature on hyperparameters importance, meta-learning, and the need for more actionable optimization strategies. Section~\ref{sec:methodology} presents the proposed MetaSHAP framework and its main components. Section~\ref{sec:experiments} reports the experimental setup and results. Finally, Section~\ref{sec:conclusion} concludes the paper and discusses directions for future research.

\section{Theoretical Background}
\label{sec:background}

\subsection{Hyperparameters Optimization (HPO)}

Hyperparameters optimization is the process of finding the set of hyperparameters values that maximize the predictive performance of a ML algorithm. Formally, let $ \mathcal{A} $ be a ML algorithm parameterized by a set of $k$ hyperparameters $ \mathpzc{h} \in \mathcal{H} $, where the hyperparameters space $ \mathcal{H} \subseteq \mathbb{R}^k $ is typically continuous, discrete, or mixed. Given a dataset $ \mathcal{D} = \{(x_i, y_i)\}_{i=1}^n $, the goal of HPO is to find an optimal configuration:
\[
\mathpzc{h}^* = \arg\max_{\mathpzc{h} \in \mathcal{H}} f(\mathcal{A}^\mathpzc{h}, \mathcal{D})
\]
where $ f(\mathcal{A}^\mathpzc{h}, \mathcal{D}) $ is a performance metric evaluating the model $\mathcal{A}$ trained with hyperparameters $ \mathpzc{h} $ on dataset $ \mathcal{D} $. Typical choices for $ f $ include accuracy, F1-score, area under the ROC curve (AUC), or negative loss functions.

Hyperparameters optimization is challenging because evaluating the performance function $f$ often requires training computationally expensive models with different HPs configurations, making exhaustive search infeasible~\cite{mgbo_sadasc}. Common strategies include grid search, random search, Bayesian optimization, evolutionary algorithms, and bandit-based methods. These approaches treat $f$ as a black-box function and focus solely on identifying well-performing hyperparameters configurations, without providing any insight into which hyperparameters are most influential or how they should be tuned (i.e., whether their values should be increased or decreased) to improve performance. This lack of transparency limits human understanding of the optimization process, impedes effective refinement and prioritization of the search space, and restricts the transfer of knowledge across different datasets and algorithms.

\subsection{Hyperparameters Importance and Interaction}
\label{hii}
In the context of hyperparameters optimization, understanding the relative contribution of each hyperparameter to model performance and the interaction among them is essential for building efficient, targeted search strategies. One principled approach to this problem is to leverage concepts from cooperative game theory, in particular the use of Shapley values~\cite{shapinml}, to quantify the importance of individual hyperparameters and their interactions. Shapley values provide a fair allocation of a total reward (or cost) among a set of players, based on their marginal contributions to all possible coalitions. When applied to HPO, we treat each hyperparameter as a player in a cooperative game, and the performance of the resulting ML model as the reward to be fairly distributed.

Formally, let $ \mathcal{H} = \{\mathpzc{h}_1, \mathpzc{h}_2, \ldots, \mathpzc{h}_k\} $ denote the set of hyperparameters. For any subset $ S \subseteq \mathcal{H} \setminus \{\mathpzc{h}_i\} $, the marginal contribution of the hyperparameter $ \mathpzc{h}_i $ is defined as the change in model performance when $ \mathpzc{h}_i $ is added to the configuration $ S $. The Shapley value $ \phi_i $ for hyperparameter $ \mathpzc{h}_i $ is given by:

\[
\phi_i = \sum_{S \subseteq \mathcal{H} \setminus \{\mathpzc{h}_i\}} \frac{|S|!(k - |S| - 1)!}{k!} \left[ f(\mathcal{A}^{(S \cup \{\mathpzc{h}_i\})} - f(\mathcal{A}^S) \right]
\]

Here, $ f(\mathcal{A}^S) $ denotes a value function representing the expected performance of the model $\mathcal{A}$ when only the hyperparameters in $ S $ are fixed (e.g., to specific values), while others are marginalized, randomized, or held at default settings. 
By applying Shapley-based analysis to surrogate performance models trained on observed pipeline results, we can estimate these importance values efficiently, even in high-dimensional hyperparameters spaces. This approach supports both post-hoc interpretation (after HPO has been run) and meta-level estimation (before launching HPO), enabling informed decisions about which HP to tune, how to restrict the search space, or which interactions merit further exploration.

\subsection{Problem Definition}

Let $ \mathcal{K} = \{(\mathcal{D}_j, \mathcal{A}_j,\mathpzc{h}_j, f_j)\}_{j=1}^M $ denote a large knowledge base comprising previously solved HPO tasks, where each entry consists of:
\begin{itemize}
    \item A dataset $ \mathcal{D}_j $, represented by a meta-feature vector $ \sigma(\mathcal{D}_j) \in \mathbb{R}^d $ encoding dataset characteristics such as number of features, class imbalance, statistical moments, etc~\cite{meta-learning}.
    \item A hyperparameter configuration $ \mathpzc{h}_j \in \mathcal{H} $.
    \item The observed performance $ f_j $ of algorithm $ \mathcal{A} $ with $ \mathpzc{h}_j $ on $ \mathcal{D}_j $.
\end{itemize}

Given a new, unseen dataset $ \mathcal{D}_{\text{new}} $ and an algorithm $ \mathcal{A} $, our objective is to predict the relative importance of each hyperparameter $ h_i \in \mathcal{H} $ \emph{before} conducting an expensive HPO procedure. Additionally, we aim to estimate the interactions between hyperparameters in order to guide the selection and range definition of the tuning space. This is accomplished by leveraging meta-learning techniques~\cite{meta-learning} to transfer knowledge from similar datasets in $\mathcal{K}$, and applying SHAP-based attribution to provide interpretable and actionable insights into hyperparameters influence.
Formally, we seek a mapping:
\[
\Phi: \left(\mathcal{D}_{\text{new}}, \mathcal{A}\right) \mapsto \{\phi_i\}_{i=1}^k
\]
where each $ \phi_i $ represents the estimated importance score of hyperparameter $ h_i $ for $\mathcal{A}$ on $ \mathcal{D}_{\text{new}} $, enabling practitioners to prioritize and understand hyperparameters interactions without costly search.


\section{Related Work}
\label{sec:related_work}

\subsection{Hyperparameters Importance Estimation}
Identifying which hyperparameters significantly affect model performance is a critical step toward efficient hyperparameters optimization. Several approaches have been proposed to estimate hyperparameters importance:

\textbf{Functional ANOVA (fANOVA)}~\cite{fanova} is among the earliest principled methods for this purpose. It models the variance of the performance function $f$ as a function of input hyperparameters, decomposing it into individual components. It efficiently quantifies which hyperparameters contribute most to performance variation, making it a valuable tool for understanding model sensitivity. While fANOVA is efficient and interpretable, it requires a sufficiently large and representative sample of the hyperparameters space and assumes an additive model, which may not always hold in practice. Moreover, fANOVA does not provide actionable insights for tunability: it does not indicate in which direction a hyperparameter should be adjusted (e.g., increased or decreased) to improve performance. Instead, it merely attributes variance, without guiding the tuning process or enabling interpretation across datasets.

\textbf{Ablation Studies} \cite{Biedenkapp2017} assess importance by measuring the performance drop when each hyperparameter is removed or fixed to a default value. Although intuitive, this method is computationally expensive and does not account for interaction effects between hyperparameters. It also lacks a solid theoretical basis for comparing importance across datasets or models.

\textbf{Shapley-based Approaches} \cite{HyperSHAP} have recently been adapted from explainable AI to the HPO setting. These approaches model hyperparameters importance by attributing the performance of a ML model to individual hyperparameters, providing theoretically grounded explanations through Shapley value decompositions. However, despite their rigorous foundation, existing works typically apply these techniques in a post-hoc manner, after the HPO process has completed, thereby limiting their capacity to inform or steer the optimization itself. Moreover, most studies operate on a per-dataset basis, treating each dataset independently without leveraging cross-dataset patterns or transferring insights from previous optimization experiences. This isolated application constrains their generalizability and leaves them ineffective in cold-start settings where no prior evaluations exist. While these approaches offer quantitative importance scores, they generally do not provide actionable tunnability information —that is, whether increasing or decreasing its value is likely to improve performance. These limitations highlight a gap in the literature: current methods focus primarily on attributing variance but fall short in delivering generalizable, and actionable tuning guidance. Our work seeks to address this gap by introducing a meta-level framework that predicts both importance and tunability patterns across datasets.

\subsection{Meta-Learning for HPO Transfer}
Meta-learning has been widely explored as a means to accelerate and warm-start HPO by leveraging historical experience across tasks~\cite{meta-learning,Chekina2011}. In the context of hyperparameter optimization, meta-learning builds models or heuristics that can generalize across datasets via meta-features—summarized statistical or structural characteristics—and maps them to relevant prior experience, such as previously evaluated algorithms or hyperparameters configurations~\cite{meta-learning}.

One widely studied family of approaches includes \textit{configuration transfer methods}, exemplified by Auto-sklearn\,2~\cite{AutoSklearn2}. These techniques retrieve datasets similar to the target based on meta-features similarity and reuse top-performing configurations from those historical tasks using BO. \textit{Ranking-based techniques}, such as those proposed by Wistuba et al.\cite{wistuba2015}, further refine this process by learning relative preferences over configurations. While effective in reducing the cost of search, these methods operate as opaque retrieval systems—they do not explain why a particular configuration is likely to succeed, nor do they offer insight into which hyperparameters drive performance. As a result, they do not readily support diagnostic or actionable tasks such as identifying influential hyperparameters, analyzing interactions, or recommending tuning directions.

Another line of research involves \textit{algorithm recommendation systems}, exemplified by AMLBID~\cite{amlbid}, which predict the best-suited learning algorithm or pipeline for a new task. These methods treat the entire algorithmic configuration as a single unit, and thus provide no transparency into the role or sensitivity of individual components or hyperparameters. This black-box treatment limits their interpretability and the practitioner’s ability to reason about model tuning\,\cite{ictai_2024}.

Despite their advances, most existing meta-learning approaches for HPO focus on \textit{what} configuration to try rather than \textit{why} and \textit{how} it works. They lack mechanisms to reason about the causal or correlational impact of individual hyperparameters in a dataset-specific manner. Consequently, they do not help practitioners prioritize which hyperparameters to tune, estimate their expected impact, or determine in which direction to adjust them. Our work addresses this critical gap by introducing a meta-interaction approach that bridges the predictive capabilities of meta-learning with actionable hyperparameters influence analysis.



 
\section{Methodology}
\label{sec:methodology}

In this section, we present MetaSHAP, a novel framework that performs dataset-aware hyperparameters importance and interaction estimation using meta-learning and Shapley-based attribution. MetaSHAP aims to provide \emph{transferable} and \emph{interpretable} insights about which hyperparameters and their exploration space are likely to influence model performance, before starting any HPO process on a new dataset.

\subsection{Overview}

MetaSHAP operates in a global HPO context, where the goal is to estimate the importance and the interactions of hyperparameters for a given algorithm on a new dataset by leveraging historical HPO data. It relies on a large-scale knowledge base $\mathcal{K}$ composed of more than 09 millions of evaluated pipelines across diverse datasets and hyperparameters configurations. Given a new dataset $\mathcal{D}_{\text{new}}$, MetaSHAP follows four main steps: (1) extract meta-features from the target dataset, (2) retrieve similar datasets, (3) train a surrogate model on relevant HPO data, and (4) apply Shapley-based attribution to derive hyperparameters importance scores.
The result is a personalized, data-driven estimation of hyperparameters importance for the given dataset and ML algorithm\,(see Algorithm\,\ref{algo1}).


\begin{algorithm}[hb]

\DontPrintSemicolon
\SetKwInOut{Input}{Input}
\SetKwInOut{Output}{Output}
\Input{
    Target dataset $\mathcal{D}_{\text{new}}$, knowledge base $\mathcal{K}$ 
    ML algorithm $\mathcal{A}$, HPs space $\mathcal{H}$, Meta-features extractor $\sigma$
    
}

\BlankLine
\textbf{Step 1: Meta-features extraction} \\
$\mathbf{m}_{\text{new}} \leftarrow \sigma(\mathcal{D}_{\text{new}})$

\BlankLine
\textbf{Step 2: Similar tasks retrieval} \\
Identify neighborhood $\mathcal{N}(\mathcal{D}_{\text{new}}) \subset \mathcal{K}$ by selecting top-$k$ datasets closest to $\mathbf{m}_{\text{new}}$ in the meta-features space

\BlankLine
\textbf{Step 3: Surrogate model training} \\
- Construct meta-dataset $\mathcal{T} = \{(\mathpzc{h}_j, f_j)\ |\ (\mathcal{D}_j, \mathcal{A}_j,\mathpzc{h}_j, f_j) \in \mathcal{N}(\mathcal{D}_{\text{new}})\}$ \\
- Train a surrogate model $\hat{f}: \mathcal{H} \to \mathbb{R}$ on $\mathcal{T}$ to approximate performance

\BlankLine
\textbf{Step 4: Hyperparameters attribution} \\
Compute marginal contributions: $\phi_i \leftarrow \text{SHAP}(\hat{f}, i),\ \forall i \in \{1, \dots, |\mathcal{H}|\}$

\BlankLine
\Return  HPs importance scores $\{\phi_i\}_{i=1}^{|\mathcal{H}|}$

\caption{MetaSHAP algorithm.}
\label{algo1}
\end{algorithm}

\subsubsection{\textbf{Step 1: Meta-feature Extraction}}

We encode the input dataset $\mathcal{D}_{new}$ as a vector of meta-features $\sigma(\mathcal{D}_{new}) \in \mathbb{R}^d$, designed to capture complementary  characteristics of the data. These meta-features serve as a compact representation that enables both dataset similarity assessment and task retrieval.  
We adopt the set of meta-features proposed in\,\cite{meta-learning,amlbid}, which can be grouped into the following categories:

\begin{itemize}
    \item \textit{General properties:} number of instances, number of features, number of classes, etc.
    \item \textit{Statistical measures:} mean, variance, skewness, kurtosis of numeric features.
    \item \textit{Information-theoretic features:} entropy, mutual information.
    \item \textit{Landmarking results:} accuracy of simple classifiers (e.g., 1-NN, decision stump).
\end{itemize}

\subsubsection{\textbf{Step 2: Similar Tasks Retrieval}}

To enable transferability, we retrieve a set of tasks $\mathcal{N}(\mathcal{D}_{\text{new}}) = \arg\min_{\mathcal{D}_j \in \mathcal{K}} \text{dist}(\sigma(\mathcal{D}_{\text{new}}), \sigma(\mathcal{D}_j))$ such that the meta-feature distance between $\sigma(\mathcal{D}_{\text{new}})$ and $\sigma(\mathcal{D}_{j})$ is minimal.

Similarity is computed using Euclidean distance.
This neighborhood serves as a proxy to model how hyperparameters behave on similar tasks, assuming that datasets with similar meta-characteristics induce similar performance landscapes in the hyperparameters space~\cite{meta-learning}.

\subsubsection{\textbf{Step 3: Surrogate Model Construction}}

Let $\mathcal{H}$ be the hyperparameters space for algorithm $\mathcal{A}$. From the neighborhood $\mathcal{N}(\mathcal{D}_{\text{new}})$, we extract a pool of evaluated configurations $\{(\mathpzc{h}_j, f_j)\}$, where $\mathpzc{h}_j \in \mathcal{H}$ and $f_j$ is the observed performance.

We then fit a surrogate regression model $\hat{f} : \mathcal{H} \to \mathbb{R}$ that predicts the performance score $f_j$ given a configuration $\mathpzc{h}_j$. This surrogate serves as a differentiable approximation of the performance function in the local neighborhood of $\mathcal{D}_{\text{new}}$.

\subsubsection{\textbf{Step 4: Shapley-based HPs Attribution}}

For each retrieved dataset we fit a surrogate model $\hat{f}_j: \mathcal{H} \to \mathbb{R} $ over observed hyperparameter-performance pairs. Hyperparameters are treated as features, and the SHAP framework is used to compute $\phi_i$ (see subsection\,\ref{hii}).

Each $\phi_i$ quantifies the average marginal contribution of hyperparameter $h_i$ to the predicted performance, marginalizing over different combinations of the other hyperparameters. This yields a ranking and magnitude of importance. In addition to individual importance, pairwise Shapley interaction indices are also computed to understand synergies or redundancies between hyperparameters\,(see subsection\,\ref{ati}).

\subsection{Extracting Actionable Tuning Insights}
\label{ati}
\begin{figure}
    \centering
    \includegraphics[width=1\linewidth]{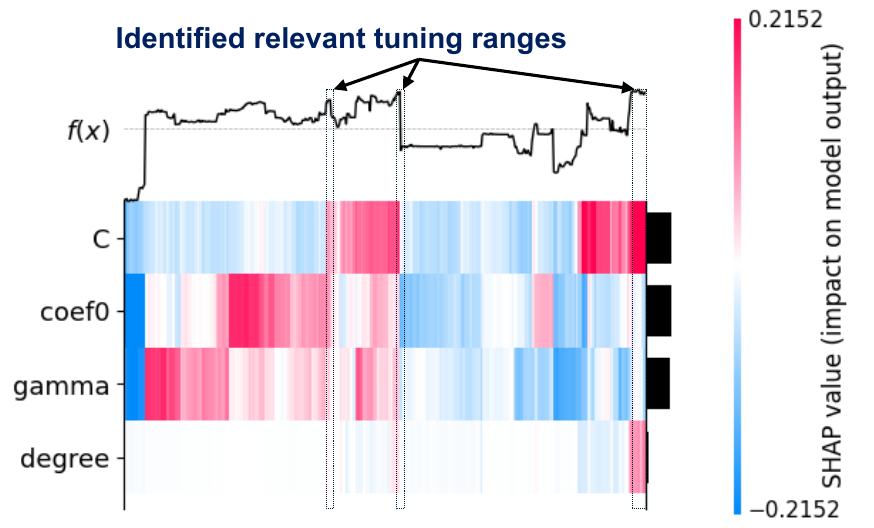}
    \caption{Heatmap displaying SHAP values for each hyperparameter, showing their impact on model output\,(SVM) on the ``ring" dataset. Red indicates a positive contribution, while blue denotes a negative one. The top curve shows the corresponding accuracy evolution. Highlighted vertical bands mark the identified relevant tuning ranges where hyperparameters exhibit the most significant influence on model behavior.}
    \label{fig:illustration}
\end{figure}

Given SHAP-based marginal contribution of hyperparameters for a given algorithm (on the neighborhood of $\mathcal{D}_{new}$), we propose a simple yet effective method to extract actionable tuning guidance. This method operates on the SHAP value distribution for the HPs and identifies the regions in the domain where the absolute impact on performance is most pronounced\,(see figure\,\ref{fig:illustration}).

Formally, for each hyperparameter $\mathpzc{h} \in \mathcal{H}$, we analyze its interaction with the model output to identify the value ranges that contribute most to performance (e.g., accuracy), as illustrated in Figure~\ref{fig:illustration}. First, we collect the corresponding SHAP values \( \phi_\mathpzc{h} \) and the actual values \( v_\mathpzc{h} \) used in the evaluated configurations. We then sort the SHAP values by \( v_\mathpzc{h} \), apply a smoothing operation (e.i., moving average), and extract the sub-ranges where the absolute smoothed SHAP values maximizes the accuracy. These intervals correspond to regions where the hyperparameter has a strong and consistent influence on model accuracy, and are thus identified as relevant tuning ranges\,(vertical bands marks in figure\,\ref{fig:illustration}).
This provides a data-driven method to identify promising sub-regions in the hyperparameters space for targeted tuning, enabling more focused and sample-efficient optimization.



\section{Evaluation}
\label{sec:experiments}

To evaluate the effectiveness of our MetaSHAP approach, we conducted experiments on a curated meta-knowledge base composed of over 09 million machine learning pipelines. We focus on the task of predicting hyperparameters importance \textit{prior} to running any optimization, using only the dataset meta-features. Our primary goal is to compare the effectiveness of this informed estimation to standard, unguided black-box optimization methods such as Bayesian Optimization.

\subsection{Meta-Knowledge Base}

To support dataset-aware hyperparameters importance analysis, we use a large-scale meta-knowledge base $ \mathcal{K} = \{(\mathcal{D}_i, \mathcal{A}_i, \mathpzc{h}_i, f_i)\}_{i=1}^M $, where each entry corresponds to a machine learning pipeline evaluated on a dataset $ \mathcal{D}_i $, using algorithm $ \mathcal{A}_i $, hyperparameter configuration $ \mathpzc{h}_i $, and resulting in performance $ f_i $.

The KB~\cite{kb} was built by running a systematic hyperparameters optimization campaign on 164 supervised classification datasets drawn from diverse sources. For each dataset, approximately 4000 distinct hyperparameters configurations per 14 algorithm were evaluated across the 164 different datasets, resulting in a total of over 09 million evaluated ML pipelines. This yields a comprehensive and diverse knowledge base that covers a wide range of data characteristics, algorithms, and tuning behaviors.
Each dataset $ \mathcal{D}_i $ is encoded by a fixed-dimensional meta-feature vector
$\sigma(\mathcal{D}_i) = [\text{n\_att}, \text{n\_inst}, \text{feat\_skewness}, \dots] \in \mathbb{R}^d$.
These meta-features capture statistical, information-theoretic, and structural properties of the data, and are used to assess dataset similarity during meta-learning.
The full list of algorithms, tuned hyperparameters, and dataset descriptions used in the KB are provided in\,\cite{kb}.

\subsection{Application to the Bayesian Optimization}

To evaluate the effectiveness and practical benefits of MetaSHAP, we evaluated its ability to guide Bayesian optimization\,(BO) by restricting the HPs search space to the most relevant dimensions and ranges. The comparison is carried out between two strategies:

\begin{itemize}
\item \textbf{Bayesian Optimization}: Standard BO with a Gaussian process surrogate model, exploring the full HPs space without prior knowledge or constraints.

\item \textbf{MetaSHAP-Guided BO}: A variant of BO in which the optimization is restricted to a subset of hyperparameters identified as important by MetaSHAP\,(subsection\,\ref{ati}).
\end{itemize}

Experiments were conducted on a subset of 06 state-of-the-art datasets, focusing on diverse classification tasks with varying characteristics. For each dataset–model pair, we run each baseline optimization strategy for a fixed budget of 30 iterations. In the MetaSHAP-guided setting, we select the 2–3 most influential hyperparameters and limit the search to their restricted subspaces as recommended by our SHAP-based interactions analysis. All other hyperparameters are held at their default values or marginalized over a narrow prior range.

\section{Results and Analysis}

\begin{figure*}[ht!]
    \centering

     \begin{subfigure}[b]{0.32\textwidth}
        \includegraphics[width=\linewidth]{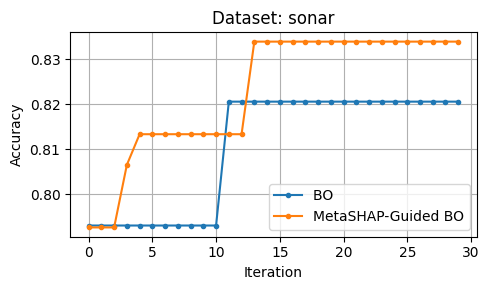}
    \end{subfigure}
    \hfill
    \begin{subfigure}[b]{0.32\textwidth}
        \includegraphics[width=\linewidth]{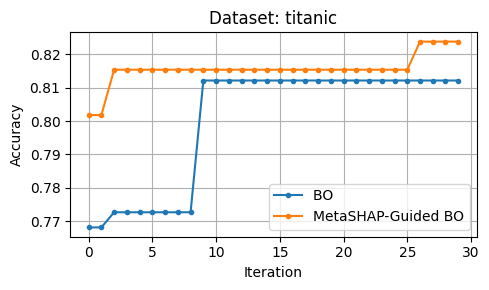}   
    \end{subfigure}
    \hfill
    \begin{subfigure}[b]{0.32\textwidth}
        \includegraphics[width=\linewidth]{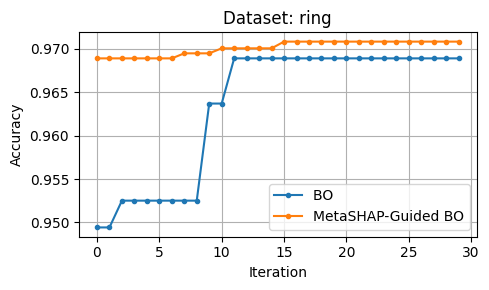}
    \end{subfigure}


    \begin{subfigure}[b]{0.32\textwidth}
        \includegraphics[width=\linewidth]{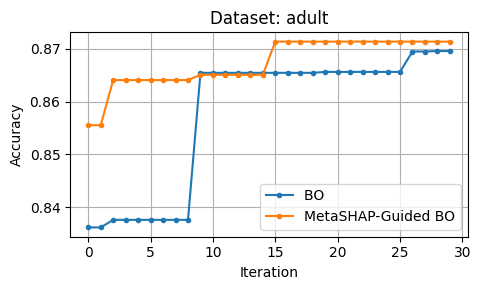}
    \end{subfigure}
    \hfill
    \begin{subfigure}[b]{0.32\textwidth}
        \includegraphics[width=\linewidth]{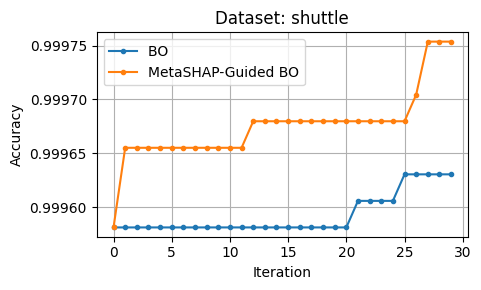}
    \end{subfigure}
    \hfill
    \begin{subfigure}[b]{0.32\textwidth}
        \includegraphics[width=\linewidth]{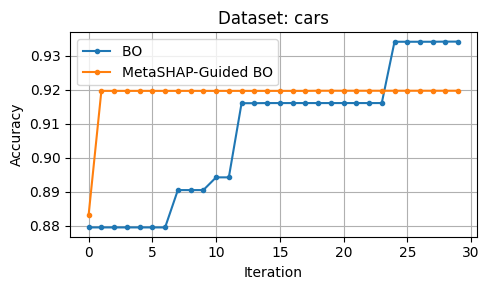}
    \end{subfigure}

    \caption{Optimization trajectories of standard Bayesian Optimization against the MetaSHAP-Guided BO.}
    \label{fig:results}
\end{figure*}

Figure~\ref{fig:results} illustrates the optimization trajectories over the 06 datasets (see Table\,\ref{tab:datasets}) using the XGBoost classifier. Each plot compares the performance of standard BO against MetaSHAP-Guided BO over 30 iterations.

\begin{table}[h!]
\centering
\caption{Description of the studied datasets.}
\label{tab:datasets}
\begin{tabular}{|l|c|c|c|}
\hline
\textbf{Dataset}            & \multicolumn{3}{c|}{\textbf{Number of}} \\ \cline{2-4}
                             & \textbf{Instances} & \textbf{Attributes} & \textbf{Classes} \\ \hline
sonar                     & 208                & 60                  & 2                \\ \hline
titanic                      & 2207               & 08                  & 2                \\ \hline
ring                         & 7400               & 20                   & 2                \\ \hline
adult                       & 48842              & 14                  & 2                \\ \hline
shuttle                     & 58000              & 9                   & 7                \\ \hline
cars                         & 392               & 8                   & 3                \\ \hline
\end{tabular}
\end{table}

A consistent trend observed across all datasets is that the MetaSHAP-guided variant reaches high-performing configurations significantly faster than the unguided baseline. For instance, in the ring dataset, MetaSHAP-Guided BO achieves near-optimal accuracy (above 0.97) from the first iteration, whereas standard BO requires around 12 iterations to converge to a similar performance level. Similarly, for the titanic dataset, the guided version stabilizes around its optimal accuracy by iteration 3, while standard BO takes more than three times as long (approximately 9 iterations) to catch up. This significant gain in efficiency highlights the benefits of initializing BO with a focused, informed search space derived from prior knowledge.

Interestingly, while guided BO generally outperforms or matches the baseline in terms of final accuracy, the cars dataset shows an exception. Here, the MetaSHAP Guided BO identifies a better configuration as early as iteration 2, while the vanilla version lags behind until around iteration 24, where it eventually surpasses the guided variant. This result suggests that the quality of MetaSHAP guidance depends on the relevance of the retrieved analog datasets from the knowledge base. If no sufficiently similar datasets are available, the risk of suboptimal guidance increases. Nonetheless, even in such cases, MetaSHAP does not severely degrade performance — demonstrating the robustness of the approach.

These findings confirm that incorporating SHAP-informed insights into the search process allows BO to prioritize promising regions of the hyperparameters space, leading to faster convergence and more interpretable tuning behavior. Moreover, MetaSHAP can serve as a complementary tool to BO or other optimization strategies, enhancing their sample efficiency without requiring major algorithmic changes.

\textbf{Interpretability of Guidance.} In addition to improving sample efficiency, the MetaSHAP-guided BO provides interpretable justifications for its restricted search. For example, for the \texttt{learning\_rate} hyperparameter in XGBoost, our method consistently suggested narrowing the range to [0.02, 0.5], a region that aligns with known best practices in the literature.

MetaSHAP typically identifies 3 to 5 hyperparameters as dominant per model-dataset pair, reducing the tuning complexity by over 50–70\% compared to the full hyperparameter set. Moreover, the extracted importance heatmaps, as shown in Figure\,\ref{fig:illustration}, reveal recurring patterns of interactions. For instance, the parameters \texttt{C}, \texttt{coef0}, and \texttt{gamma} in the SVM model tend to exhibit synergistic effects, whereas parameters such as \texttt{degree} often show negligible contributions on the ``ring" dataset. High values of \texttt{C} consistently lead to strong SVM performance. In contrast, \texttt{coef0} does not exhibit a clear trend across its value range in relation to model performance. These insights are human-interpretable and can support manual or hybrid tuning strategies in practical applications.


\section{Conclusion}
\label{sec:conclusion}

This paper introduces MetaSHAP, a semi-automated explainable framework that bridges meta-learning and Shapley-based importance analysis to guide hyperparameters optimization in a transparent and data-efficient manner. Unlike conventional approaches that treat hyperparameters tuning as a purely black-box optimization problem, MetaSHAP provides actionable insights by identifying the most influential hyperparameters and quantifying their interactions. Leveraging a large-scale meta-knowledge base of over 09 million evaluated ML pipelines, the framework retrieve and transfer knowledge in the form of hyperparameters importance rankings and tuning ranges. Our experimental results demonstrate that MetaSHAP-guided Bayesian Optimization consistently outperforms traditional BO in both accuracy and convergence speed, while significantly reducing the dimensionality of the search space. The interpretability of the output further empowers practitioners to make informed decisions, streamline tuning processes, and better understand model behavior. These findings highlight the potential of combining principled attribution methods with prior knowledge for improving the efficiency and transparency of AutoML workflows. Looking forward, MetaSHAP opens promising directions for generalizing this approach to regression tasks, deep neural networks, and other forms of parameterized learning algorithms, ultimately contributing to more intelligent and explainable machine learning systems.

\bibliographystyle{IEEEtran}
\bibliography{bibliography}

\end{document}